\documentclass[10pt,twocolumn,letterpaper]{article}

\usepackage{iccv}
\usepackage{times}
\usepackage{epsfig}
\usepackage{graphicx}
\usepackage{amsmath}
\usepackage{amssymb}
\newtheorem{theorem}{Remark}
\usepackage{multicol}
\usepackage{multirow}
\usepackage{booktabs}
\usepackage{soul}
\usepackage{caption}
\usepackage[pagebackref=true,breaklinks=true,letterpaper=true,colorlinks,bookmarks=false]{hyperref}

\iccvfinalcopy % *** Uncomment this line for the final submission

\def\iccvPaperID{2894} % *** Enter the ICCV Paper ID here

\ificcvfinal\fi

\begin{document}

%%%%%%%%% TITLE
\title{Parametric Contrastive Learning}
\author{
	Jiequan Cui $^{1}$ \quad 
	Zhisheng Zhong $^{1}$ \quad
	Shu Liu $^{2}$ \quad
	Bei Yu $^{1,2}$ \quad
	Jiaya Jia $^{1,2}$ \\
	$^{1}$The Chinese University of Hong Kong \hspace{1cm} $^{2}$SmartMore \hspace{1cm} 
    \vspace{.7em}\\
	{\tt\small \{jqcui, zszhong21, byu, leojia\}@cse.cuhk.edu.hk, \{liushuhust\}@gmail.com}
}

\maketitle
\ificcvfinal\thispagestyle{empty}\fi

\begin{abstract}
	In this paper, we propose Parametric Contrastive Learning (PaCo) to tackle long-tailed recognition. Based on theoretical analysis, we observe supervised contrastive loss tends to bias on high-frequency classes and thus increases the difficulty of imbalanced learning. We introduce a set of parametric class-wise learnable centers to rebalance from an optimization perspective. Further, we analyze our PaCo loss under a balanced setting. Our analysis demonstrates that PaCo can adaptively enhance the intensity of pushing samples of the same class close as more samples are pulled together with their corresponding centers and benefit hard example learning. Experiments on long-tailed CIFAR, ImageNet, Places, and iNaturalist 2018 manifest the new state-of-the-art for long-tailed recognition. On full ImageNet, models trained with PaCo loss surpass supervised contrastive learning across various ResNet backbones, {\it e.g.}, our ResNet-200 achieves 81.8\% top-1 accuracy. Our code is available at \url{https://github.com/dvlab-research/Parametric-Contrastive-Learning}.
\end{abstract}

\section{Introduction}
Convolutional neural networks (CNNs) have achieved great success in various tasks, including image classification \cite{he2016deep,vggnet}, object detection \cite{DBLP:conf/cvpr/LinDGHHB17, DBLP:conf/cvpr/LiuQQSJ18} and semantic segmentation \cite{DBLP:conf/cvpr/ZhaoSQWJ17}. With neural network search \cite{DBLP:conf/cvpr/ZophVSL18, DBLP:conf/iclr/LiuSY19, DBLP:conf/cvpr/TanCPVSHL19, DBLP:conf/iccv/CuiCLLSJ19, DBLP:conf/iclr/CaiGWZH20}, performance of CNNs further boosts. Impressive progress highly depends on large-scale and high-quality datasets, such as ImageNet \cite{imagenet}, MS COCO \cite{coco} and Places \cite{zhou2017places}. When dealing with real-world applications, generally we face the long-tailed distribution problem -- a few classes contain many instances, while most classes contain only a few instances. Learning in such an imbalanced setting is challenging as the low-frequency classes can be easily overwhelmed by high-frequency ones. Without considering this situation, CNNs will suffer from significant performance degradation. 

\begin{figure}[tb!]
	\begin{center}
		\includegraphics[width=1.06\linewidth]{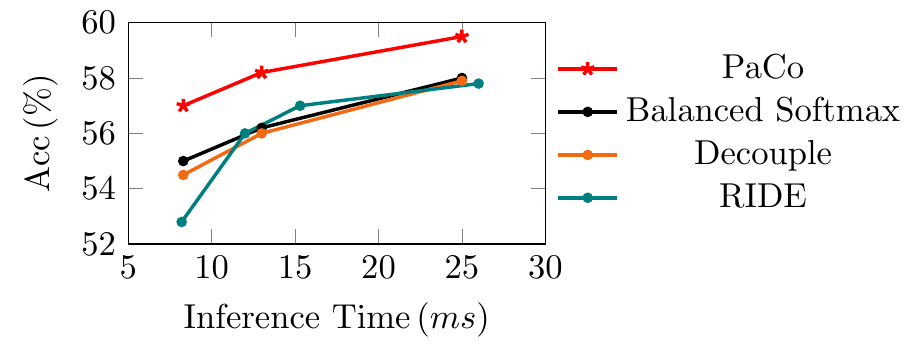}
		\caption{Comparison with state-of-the-arts on ImageNet-LT \cite{Liu_2019_CVPR}. Inference time is calculated with a batch of 64 images on Nvidia GeForce 2080Ti GPU. The same experimental setting is adopted for comparison. ResNet-50, ResNeXt-50, and ResNeXt101 are used for Balanced Softmax \cite{DBLP:conf/nips/RenYSMZYL20}, Decouple \cite{DBLP:conf/iclr/KangXRYGFK20}, and PaCo. For RIDE \cite{DBLP:journals/corr/abs-2010-01809}, various numbers of expert with RIDEResNet and RIDEResNeXt are adopted. PaCo significantly outperforms recent SOTA. Detailed numbers for RIDE is in the supplementary file.}
		\label{fig:PaCo_comparison}
		\vspace{-0.1in}
	\end{center}
\end{figure}

Contrastive learning \cite{DBLP:conf/icml/ChenK0H20, DBLP:conf/cvpr/He0WXG20, DBLP:journals/corr/abs-2003-04297, DBLP:conf/nips/GrillSATRBDPGAP20, DBLP:conf/nips/CaronMMGBJ20} is a major research topic due to its success in self-supervised representation learning. Khosla \etal.~\cite{DBLP:conf/nips/KhoslaTWSTIMLK20} extend non-parametric contrastive loss into non-parametric supervised contrastive loss by leveraging label information, which trains representation in the first stage and learns the linear classifier with the fixed backbone in the second stage. Though supervised contrastive learning works well in a balanced setting, for imbalanced datasets, our theoretical analysis shows that high-frequency classes will have a higher lower bound of loss and contribute much higher importance than low-frequency classes when equipping it in training.

\begin{figure}[t]
	\begin{center}
		\includegraphics[width=0.98\linewidth]{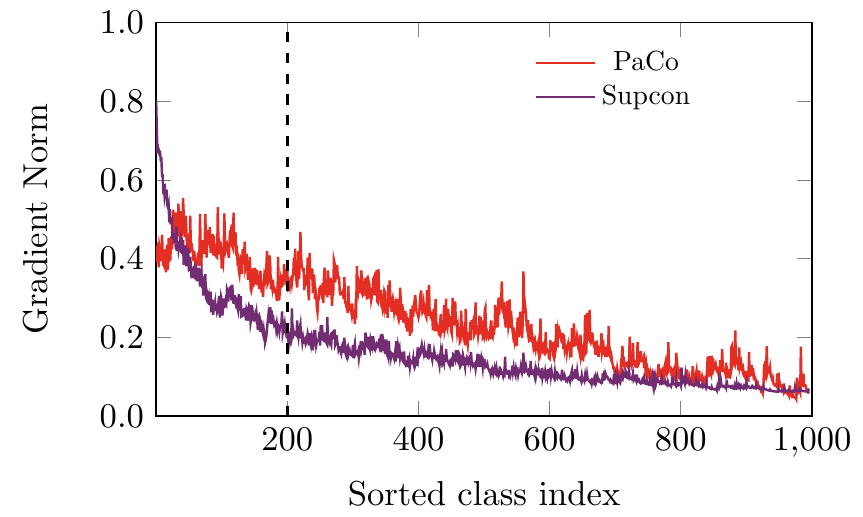}
		\caption{Rebalance in contrastive learning. We collect the average L2 norm of the gradient of weights in the last classifier layer on ImageNet-LT. Category indices are
			sorted by their image counts. The gradient norm varying from the most frequent class to the least one is steep for supervised contrastive learning \cite{DBLP:conf/nips/KhoslaTWSTIMLK20}. In particular, the gradient norm dramatically decreases for the top 200 most frequent classes. Trained with PaCo, the gradient norm is better balanced.}
		\label{fig:gradient}
	\end{center}
\end{figure}

This phenomenon leads to model bias on high-frequency classes and increases the difficulty of imbalanced learning.
As shown in Fig.~\ref{fig:gradient}, when the model is trained with supervised contrastive loss on ImageNet-LT, the gradient norm varying from the most frequent class to the least one is rather steep. In particular, the gradient norm dramatically decreases for the top 200 most frequent classes.    

Previous work \cite{DBLP:journals/nn/BudaMM18,DBLP:conf/cvpr/HuangLLT16,DBLP:conf/cvpr/CuiJLSB19, he2009learning,chawla2002smote, shen2016relay, DBLP:conf/nips/RenYSMZYL20, DBLP:conf/iclr/KangXRYGFK20, DBLP:journals/corr/abs-2010-01809, DBLP:journals/corr/abs-2101-10633, DBLP:conf/nips/TangHZ20, DBLP:journals/corr/abs-2101-10633, Zhong_2021_CVPR} explored rebalancing in traditional supervised cross-entropy learning. In this paper, we tackle the above mentioned imbalance issue in supervised contrastive learning and make use of contrastive learning for long-tailed recognition. To our knowledge, it is the first attempt of using rebalance in contrastive learning. 

To rebalance in supervised contrastive learning, we introduce a set of parametric class-wise learnable centers into supervised contrastive learning. We name our algorithm \textbf{Parametric Contrastive Learning (PaCo)} shown in Fig.~\ref{fig:PaCo_framework}. With such a simple and yet effective operation, we theoretically prove that the optimal values for the probability that two samples are a true positive pair (belonging to the same class), varying from the most frequent class to the least frequent class, are more balanced. Thus their lower bound of loss values are better organized. This phenomenon means the model takes more care of low-frequency classes, making the PaCo loss benefit imbalanced learning. Fig.~\ref{fig:gradient} shows that, with our PaCo loss in training, gradient norm varying from the most frequent class to the least one are moderated better than supervised contrastive learning, which matches our analysis. 

Further, we analyze the PaCo loss under a balanced setting. Our analysis demonstrates that with more samples clustered around their corresponding centers in training, the PaCo loss increases the intensity of pushing samples of the same class close, which benefits hard examples learning.

Finally, we conduct experiments on long-tailed version of CIFAR \cite{cb-focal, DBLP:conf/nips/CaoWGAM19}, ImageNet \cite{Liu_2019_CVPR}, Places \cite{Liu_2019_CVPR} and iNaturalist 2018 \cite{van2018inaturalist}. Experimental results show that we create a new record for long-tailed recognition. We also conduct experiments on full ImageNet \cite{imagenet} and CIFAR \cite{krizhevsky2009learning}. ResNet models trained with PaCo also outperform the ones by supervised contrastive learning on such balanced datasets. Our key contributions are as follows.
\begin{itemize}
	\item We identify the shortcoming of supervised contrastive learning under an imbalanced setting -- it tends to bias towards high-frequency classes.
	\vspace{-5pt}
	\item We extend supervised contrastive loss to the PaCo loss, which is more friendly to imbalance learning, by introducing a set of parametric class-wise learnable centers.
	\vspace{-15pt}
	\item Equipped with the PaCo loss, we create new record across various benchmarks for long-tailed recognition.
	Moreover, experimental results on full ImageNet and CIFAR validate the effectiveness of PaCo under a balanced setting.
\end{itemize}

\section{Related Work}
\paragraph{Re-sampling/re-weighting.}
The most classical way to deal with long-tailed datasets is to over-sample low-frequency class images \cite{shen2016relay, zhong2016towards, buda2018systematic, byrd2019effect}
or under-sample high-frequency class images \cite{he2009learning, japkowicz2002class,buda2018systematic}. However, Oversampling can suffer from heavy over-fitting to low-frequency
classes especially on small datasets. For under-sampling, discarding a large portion of high-frequency class data inevitably causes degradation of the generalization ability of CNNs. Re-weighting \cite{huang2016learning, huang2019deep,wang2017learning,ren2018learning,shu2019meta,jamal2020rethinking} the loss functions is an alternative way to rebalance by either enlarging weights on more challenging and sparse classes or randomly ignoring gradients from high-frequency classes \cite{tan2020eql}. However, with large-scale data, re-weighting makes CNNs difficult to optimize during training \cite{huang2016learning, huang2019deep}.

\paragraph{One/two-stage Methods.}
Since deferred re-weighting and re-sampling were proposed by Cao \etal.~\cite{DBLP:conf/nips/CaoWGAM19}, Kang \etal.~\cite{DBLP:conf/iclr/KangXRYGFK20} and Zhou \etal.~\cite{zhou2019bbn} observed re-weighting or re-sampling strategies could benefit classifier learning while hurting representation learning. Kang \etal.~\cite{DBLP:conf/iclr/KangXRYGFK20} proposed to decompose representation and classifier learning. It first trains the CNNs with uniform sampling, and then fine-tune the classifier with class-balanced sampling while keeping parameters of representation learning fixed. Zhou \etal.~\cite{zhou2019bbn} proposed one cumulative learning strategy, with which they bridge representation learning and classifier re-balancing.

The two-stage design is not for end-to-end frameworks.
Tang \etal.~\cite{DBLP:conf/nips/TangHZ20} analyzed the reason from the perspective of causal graph and concluded that the bad momentum causal effects played a vital role.
Cui \etal.~\cite{DBLP:journals/corr/abs-2101-10633} proposed residual learning mechanism to address this issue.

\paragraph{Non-parametric Contrastive Loss.}
\label{Sec:contrastive_learning}
Contrastive learning \cite{DBLP:conf/icml/ChenK0H20, DBLP:conf/cvpr/He0WXG20, DBLP:journals/corr/abs-2003-04297, DBLP:conf/nips/GrillSATRBDPGAP20, DBLP:conf/nips/CaronMMGBJ20} is a framework that learns similar/dissimilar representations from data that are organized into similar/dissimilar pairs. An effective contrastive loss function, called InfoNCE \cite{DBLP:journals/corr/abs-1807-03748}, is
\begin{equation}
\mathcal{L}_{q, k^+, \{k^-\}} = -\log \frac{\exp(q{\cdot}k^+ / \tau)}{\exp(q{\cdot}k^+ / \tau) + {\displaystyle\sum_{k^-}}\exp(q{\cdot}k^-  / \tau)},
\label{eq:infonce}
\end{equation}
where $q$ is a query representation, $k^+$ is for the positive (similar) key sample, and $\{k^-\}$ denotes negative (dissimilar) key samples. $\tau$ is a temperature hyper-parameter. In the {instance discrimination} pretext task \cite{DBLP:conf/cvpr/WuXYL18} for self-supervised learning, a query and a key form a positive pair if they are data-augmented versions of the same image. It forms a negative pair otherwise. 

Traditional cross-entropy with linear fc layer weight $w$ and true label $y$ among $n$ classes is expressed as 
\begin{equation}
\mathcal{L}_{cross-entropy} = -\log \frac{\exp(q{\cdot} w_{y})}{\sum_{i=1}^{n} \exp(q{\cdot}w_{i})}.
\label{eq:cross-entropy}
\end{equation}

Compared to it, InfoNCE does not get involved with parametric learnable parameters.
To distinguish our proposed parametric contrastive learning from previous ones, we treat the InfoNCE as a non-parametric contrastive loss following \cite{wu2018unsupervised}.

Chen \etal.~\cite{DBLP:conf/icml/ChenK0H20} used self-supervised contrastive learning SimCLR to first match the performance of a supervised ResNet-50 with only a linear classifier trained on self-supervised representation on full ImageNet.
He \etal.~\cite{DBLP:conf/cvpr/He0WXG20} proposed MoCo and Chen \etal~\cite{DBLP:journals/corr/abs-2003-04297} extended MoCo to MoCo v2, with which small batch size training can also achieve competitive results on full ImageNet \cite{imagenet}.
In addition, many other methods \cite{DBLP:conf/nips/GrillSATRBDPGAP20, DBLP:conf/nips/CaronMMGBJ20} are also proposed to further boost performance. 

\section{Parametric Contrastive Learning}
\begin{figure*}[t]
	\begin{minipage}{.616\linewidth}
		\centering
		\includegraphics[width=1.00\textwidth]{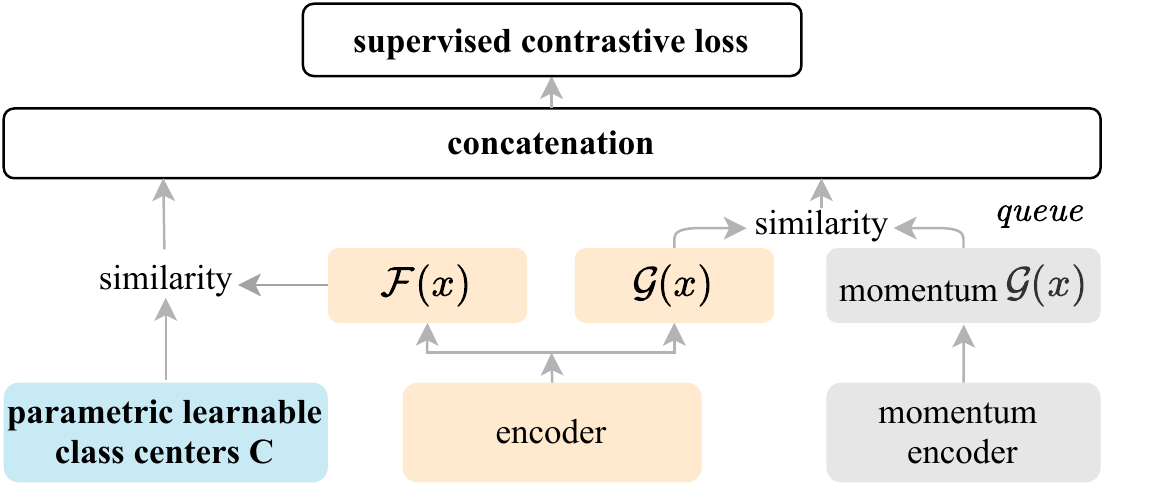}
		\captionof{figure}{Parametric contrastive learning (PaCo). We introduce a set of parametric class-wise learnable centers for rebalancing in contrastive learning. More analysis is in Section \ref{Sec:PaCo} for PaCo.}
		~\label{fig:PaCo_framework}
	\end{minipage}
	\vspace{.1in}
	\begin{minipage}{.38\linewidth}
		\centering
		\includegraphics[width=1.00\linewidth]{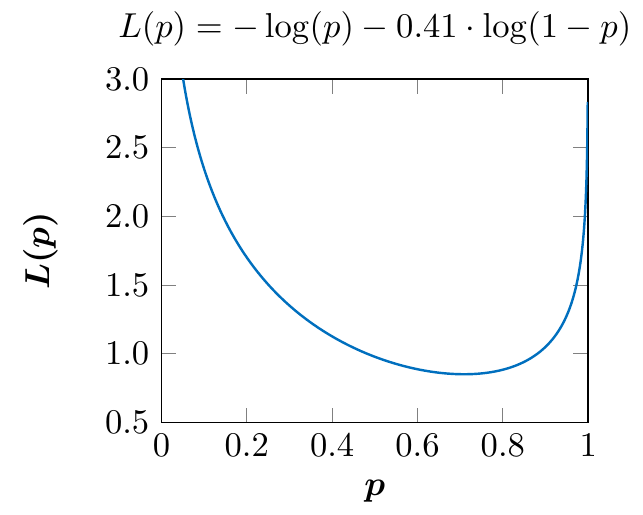}
		\captionof{figure}{Curve for $\mathcal{L}_{extra}$.}
		\label{fig:psum}
	\end{minipage}
\end{figure*}

\subsection{Supervised Contrastive Learning}
\label{Sec:supcon}

Khosla \etal~\cite{DBLP:conf/nips/KhoslaTWSTIMLK20} extended the self-supervised contrastive loss with label information into supervised contrastive loss. Here we present it in the framework of MoCo \cite{DBLP:conf/cvpr/He0WXG20, DBLP:journals/corr/abs-2003-04297} as
\begin{equation}
\mathcal{L}_{i} \!=\! -\sum_{ z_{+} \in P(i)} \log \frac{ \exp({z_{+} \cdot T(x_{i})}) }{\sum_{z_{k} \in A(i)} \exp({z_{k} \cdot T(x_{i})})}. \label{eq:supcon} \\
\end{equation}
MoCo framework \cite{DBLP:conf/cvpr/He0WXG20, DBLP:journals/corr/abs-2003-04297} consists of two networks with the same structure, {\it i.e.}, { \it query network} and {\it key network}. The key network is driven by a momentum update with the query network in training. For each network, it usually contains one encoder CNN and one two-layer MLP transform.

During training, for one two-viewed image batch $B=(B_{v1}, B_{v2})$ and label $y$, $B_{v1}$ and $B_{v2}$ are fed into the {query network} and {key network} respectively and we denote their outputs as $Z_{v1}$ and $Z_{v2}$.
Especially, $Z_{v2}$ is used to update the momentum {\it queue}. 

In Eq.~\eqref{eq:supcon}, $x_{i}$ is the representation for image $X_{i}$ in $B_{v1}$ obtained by the encoder of {query network}. The transform $T(\cdot)$ also belongs to the {query network}. We write
\begin{align*}
A(i) ={}& \{z_{k} \in queue \cup Z_{v1} \cup Z_{v2}\} \backslash \{ z_{k} \in Z_{v1}: k=i\}, \\
P(i) ={}& \{z_{k} \in A(i): y_{k} = y_{i}\}.
\end{align*}

In implementation, the loss is usually scaled by $\cfrac{1}{|P(i)|}$ and a temperature $\tau$ is applied like in Eq.~\eqref{eq:infonce}.
Different from self-supervised contrastive loss, which treats query and key as a positive pair if they are the data-augmented version of the same image, supervised contrastive loss treats them as one positive pair if they belong to the same class. 

\subsection{Theoretical Motivation}
\paragraph{Analysis of Supervised Contrastive Learning.}
Khosla \etal~\cite{DBLP:conf/nips/KhoslaTWSTIMLK20} introduced supervised contrastive learning to encourage more compact representation. We observe that it is not directly applicable to long-tailed recognition.
As shown in Table~\ref{tab:motivation}, the performance significantly decreases compared with traditional supervised cross-entropy.
From an optimization point of view, supervised contrastive loss concentrates more on high-frequency classes than low-frequency ones, which is unfriendly for imbalanced learning.

\begin{table}[tb!]
	\small
	\centering
	\caption{Top-1 accuracy (\%) of supervised contrastive learning on ImageNet-LT with ResNet-50. Implementation details are in supplementary file. \dag represents that the model is trained with PaCo loss without center learning rebalance.}
	\vspace{0pt}
	{
		\begin{tabular}{lcccc}
			\toprule
			Method &Many &Medium &Few &All \\
			\midrule
			Cross-Entropy &67.5 &42.6 &13.7 &48.4\\
			SupCon        &53.4 &2.9 &0 &22.0\\
			PaCo (ours) \dag    &69.6 &45.8 &16.0 &51.0\\
			\bottomrule
		\end{tabular}
	}
	\label{tab:motivation}
\end{table}

\begin{theorem}
	\label{thm:01_supcon}
	\normalfont{(Optimal value for supervised contrastive learning).}
	When supervised contrastive loss converges, the optimal value for the probability that two samples are a true positive pair with label $y$ is $\cfrac{1}{K_{y}}$, where, $q(y)$ is the frequency of class $y$ over the whole dataset, {\it queue} is the momentum  {\it queue} in MoCo \cite{DBLP:conf/cvpr/He0WXG20, DBLP:journals/corr/abs-2003-04297} and $K_{y} \approx \mathtt{length}(queue) \cdot q(y)$.
	%\textit{Proof~~} See supplementary material.
\end{theorem}    

\paragraph{Interpretation.} As indicated by Remark \ref{thm:01_supcon}, high-frequency classes have a higher lower bound of loss value and contribute much higher importance than low-frequency classes in training. Thus the training process can be dominated by high-frequency classes. To handle this issue, we introduce a set of parametric class-wise learnable centers for rebalancing in contrastive learning.

\subsection{Rebalance in Contrastive Learning}
\label{Sec:PaCo}
As described in Fig.~\ref{fig:PaCo_framework}, we introduce a set of parametric class-wise learnable centers $\mathbf{C} =\{c_{1}, c_{2},..., c_{n} \}$ into the original supervised contrastive learning, and call this new form {\bf Parametric Contrastive Learning (PaCo)}. Correspondingly, the loss function is changed to
\begin{equation}
\mathcal{L}_{i} \!=\! \sum_{ z_{+} \in P(i) \cup \{c_{y}\} } \!-w(z_{+}) \log \frac{ \exp({z_{+} \cdot T(x_{i})}) }{\sum_{z_{k} \in A(i) \cup \mathbf{C}} \exp({z_{k} \cdot T(x_{i})})},\label{eq:PaCo}
\end{equation}
\text{where}
\iffalse
\begin{gather*}
w(x_{+})\!=\!\left\{
\begin{aligned}
\alpha, \!\quad\! z_{+} \!\in\! P(i) \\
1.0, \!\quad\! z_{+} \!\in\! \{c_{y} \}
\end{aligned}
\right.,
\!\quad\!
T(x_{i})\!=\!\left\{
\begin{aligned}
\mathcal{G}(x_{i}), \!\quad\! z_{+} \!\in\! P(i) \\
\mathcal{F}(x_{i}), \!\quad\! z_{+} \!\in\! \{c_{y} \}
\end{aligned}
\right.
\end{gather*}
\fi
\begin{equation}
w(z_{+})\!=\!\left\{
\begin{array}{lr}
\alpha, \!\quad\! z_{+} \!\in\! P(i) &  \\
1.0, \!\quad\! z_{+} \!\in\! \{c_{y} \} & \\
\end{array}
\right.
\nonumber
\end{equation}
and
\begin{equation}
z \cdot T(x_{i})\!=\!\left\{
\begin{array}{lr}
z \cdot \mathcal{G}(x_{i}), \!\quad\! z \!\in\! A(i) &  \\
z \cdot \mathcal{F}(x_{i}), \!\quad\! z \!\in\! \mathbf{C}. & \\
\end{array}
\right.
\nonumber
\end{equation}

Following Chen \etal~\cite{DBLP:journals/corr/abs-2003-04297}, the transform $\mathcal{G}(\cdot)$ is a two-layer MLP while $\mathcal{F}(\cdot)$ is the identity mapping, {\textit i.e.}, $\mathcal{F}(x)=x$. $\alpha$ is one hyper-parameter in $(0,1)$. $P(i)$ and $A(i)$ are the same with supervised contrastive learning in Eq.~\eqref{eq:supcon}. In implementation, the loss is scaled by $\cfrac{1}{\sum_{ z_{+} \in P(i) \cup \{ c_{y} \} } w(z_{+})}$ and a temperature $\tau$ is applied like that in Eq.~\eqref{eq:supcon}.

\begin{theorem}
	\label{thm:02_pcl}
	\normalfont{(Optimal value for parametric contrastive learning)~~}
	When parametric contrastive loss converges, the optimal value for the probability that two samples are a true positive pair with label $y$ is $\cfrac{\alpha}{1+\alpha \cdot K_{y}}$ and the optimal value for the probability that a sample is closest to its corresponding center $c_{y}$ among $\mathbf{C}$ is $\cfrac{1}{1+\alpha \cdot K_{y}}$, where $q(y)$ is the frequency of class $y$ over the whole dataset, {\it queue} is the momentum  {queue} in MoCo \cite{DBLP:conf/cvpr/He0WXG20, DBLP:journals/corr/abs-2003-04297} and $K_{y} \approx \mathtt{length}(queue) \cdot q(y)$.
	%\textit{Proof~~} See the Supplementary Material.
\end{theorem}  

\paragraph{Interpretation.} Suppose the most frequent class $y_{h}$ has $K_{y_{h}} \approx q(y_{h}) \cdot \mathtt{length}(queue)$ and the least frequent class $y_{t}$ has $K_{y_{t}} \approx q(y_{t}) \cdot \mathtt{length}(queue)$. As indicated by Remarks \ref{thm:02_pcl} and \ref{thm:01_supcon}, the optimal value for the probability that two samples are a true positive pair varying from the most frequent class to the least one is rebalanced from $\cfrac{1}{K_{y_{h}}}$ $\longrightarrow$ $\cfrac{1}{K_{y_{t}}}$ to $\cfrac{1}{\frac{1}{\alpha} + K_{y_{h}}}$ $\longrightarrow$ $\cfrac{1}{\frac{1}{\alpha} + K_{y_{t}}}$. The smaller $\alpha$ is, the more uniform the optimal value from the most frequent class to the least one is, friendly to low-frequency classes learning. 

However, when $\alpha$ decreases, the intensity of contrast among samples becomes weaker, the intensity of contrast between samples and centers is stronger. The whole loss becomes closer to supervised cross-entropy. To make good use of contrastive learning and rebalance at the same time, we observe that $\alpha$=0.05 is a reasonable choice.

\subsection{PaCo under Balanced Setting}
\label{sec:PaCo_balance}
For balanced datasets, all classes have the same frequency, {\it i.e.}, $q^{*}$=$q(y_{i})$=$q(y_{j})$ and $K^{*}$=$K_{y_{i}}$=$K_{y_{j}}$ for any class $y_{i}$ and class $y_{j}$. In this case, PaCo reduces to an improved version of multi-task with the weighted sum of supervised cross-entropy loss and supervised contrastive loss. The connection between PaCo and multi-task is
\begin{equation*}
ExpSum = \sum_{c_{k} \in \mathbf{C}} \exp(c_{k} \!\cdot\! \mathcal{F}(x_{i})) + \sum_{z_{k} \in A(i)} \exp(z_{k} \cdot \mathcal{G}(x_{i})).
\end{equation*}

We also write the PaCo loss as
\begin{align*}
\mathcal{L}_{i} 
={}& \sum_{ z_{+} \in P(i) \cup \{c_{y}\} } \!-w(z_{+}) \log \frac{ \exp({z_{+} \cdot T(x_{i})}) }{\sum_{z_{k} \in A(i) \cup \mathbf{C}} \exp({z_{k} \cdot T(x_{i})})} \\
={}& -\log \frac{\exp(c_{y} \cdot \mathcal{F}(x_{i}) )}{ExpSum} - \alpha\sum_{z_{+} \in P(i) }\log \frac{\exp(z_{+} \cdot \mathcal{G}(x_{i}) )}{ExpSum}\\
={}& \mathcal{L}_{sup} \!+\! \alpha  \mathcal{L}_{supcon} \!-\! \left( \log P_{sup} \!+\! \alpha K^{*}  \log P_{supcon} \right)  \\
={}& \mathcal{L}_{sup} \!+\! \alpha \mathcal{L}_{supcon} \!-\! \left( \log P_{sup} \!+\! \alpha K^{*}  \log (1-P_{sup}) \right), \\
%\label{eq:pcl_multitask}
\notag
\end{align*}
\begin{equation}
\text{where}\quad \  \left\{
\begin{aligned}
P_{sup} &= \frac{\sum_{c_{k} \in \mathbf{C}} \exp(c_{k} \!\cdot\! \mathcal{F}(x_{i}))}{ExpSum}; \\
P_{supcon} &= \frac{\sum_{z_{k} \in A(i)} \exp(z_{k} \cdot \mathcal{G}(x_{i}))}{ExpSum}. \label{eq:pcl_multitask} \\
\end{aligned}
\right.
\end{equation}

Multi-task learning combines supervised cross-entropy loss and supervised contrastive loss with a fixed weighted scalar. When these two losses conflict,  training can suffer from slow convergence or sub-optimization. Our PaCo contrarily adjusts the intensity of supervised cross-entropy loss and supervised contrastive loss in an adaptive way and potentially avoids conflict as analyzed in the following.

\subsubsection{Analysis of PaCo under Balanced Setting}
As indicated by Eq.~\eqref{eq:pcl_multitask}, compared with multi-task, PaCo has an additional loss item of
\begin{equation}
\mathcal{L}_{extra}=-\log(P_{sup}) - \alpha K^{*} \log (1-P_{sup}). \label{eq:PaCo_extra}
\end{equation}

Here, we take full ImageNet as an example, {\it i.e.}, $q^{*}=0.001$, $\mathtt{length}(queue)=8192$, $\alpha=0.05$, and $\alpha K^{*}=0.41$.
Then the function curve for $\mathcal{L}_{extra}$ is shown in Fig.~\ref{fig:psum}.
With $P_{sup}$ increases from 0 to 1.0, the function value decreases until $P_{sup}=0.71$ and then goes up, which implies $\mathcal{L}_{extra}$ obtains the smallest loss value when $P_{sup}=0.71$. Note that, when the whole PaCo loss in Eq.~\eqref{eq:PaCo} achieves the optimal solution, $P_{sup}=0.71$ still holds as demonstrated in Remark \ref{thm:02_pcl}. With $P_{sup}$ increases during training, we analyze how it affects the intensity of supervised contrastive loss and supervised cross-entropy loss in the following.

\paragraph{Adaptive Weighting between $\mathcal{L}_{sup}$ and $\mathcal{L}_{supcon}$.} 
Note that the optimal value is 0.035 for the probability that two samples are a true positive pair with label $y$ as indicated by Remark \ref{thm:02_pcl}. We suppose $p_{l},p_{h}\in (0, 0.71)$ and $p_{l}$ $\textless$ $p_{h}$. To achieve the optimal value, when $P_{sup}$=$p$, the supervised contrastive loss value $\mathcal{L}_{supcon}$ decreases as shown in Eq.~\eqref{eq:PaCo_adaptive}.

\begin{figure}[htb!]
	\begin{equation}
	\begin{aligned}
	\mathcal{L}_{supcon} 
	={}& -\sum_{z_{+} \in P(i)} \log \frac{\exp(z_{+} \cdot \mathcal{G}(x_{i}) )}{\sum_{z_{k} \in A(i)} \exp(z_{k} \cdot \mathcal{G}(x_{i})) } \\
	={}& -\sum_{z_{+} \in P(i)} \log \frac{ \frac{\exp(z_{+} \cdot \mathcal{G}(x_{i}) )}{ExpSum}}{\frac{\sum_{z_{k} \in A(i)} \exp(z_{k} \cdot \mathcal{G}(x_{i}))}{ExpSum}} \\
	={}& -K^{*} \log \frac{0.035}{1-p}. \label{eq:PaCo_adaptive} 
	\end{aligned}
	\end{equation}
\end{figure}

Here $P_{sup}$ increases from $p_{l}$ to $p_{h}$, $\mathcal{L}_{supcon}$ decreases to a much smaller loss value to achieve the optimal solution, which implies the need to make two different class samples much more discriminative, {\it i.e.}, increasing inter-class margins. Thus the intensity of supervised contrastive loss enlarges.

An intuition is that as $P_{sup}$ increases, more samples are pulled together with their corresponding centers.
Along with stronger intensity of supervised contrastive loss at that time, it is more likely to push hard examples close to those samples that are already around right centers.

\subsection{Center Learning Rebalance}
PaCo balances the contrastive learning (for moderating contrast among samples). However the center learning also needs to be balanced, which has been explored in \cite{DBLP:journals/nn/BudaMM18,DBLP:conf/cvpr/HuangLLT16,DBLP:conf/cvpr/CuiJLSB19, he2009learning,chawla2002smote, shen2016relay, DBLP:conf/nips/RenYSMZYL20, DBLP:conf/iclr/KangXRYGFK20, DBLP:journals/corr/abs-2010-01809, DBLP:journals/corr/abs-2101-10633, DBLP:conf/nips/TangHZ20,duggal2020elf, Zhong_2021_CVPR}. We incorporate Balanced Softmax \cite{DBLP:conf/nips/RenYSMZYL20} into the center learning. Then the PaCo loss is changed from Eq.~\eqref{eq:PaCo} to
\begin{equation}
\mathcal{L}_{i} \!=\! \sum_{ z_{+} \in P(i) \cup \{c_{y}\} } \!-w(z_{+}) \log \frac{ \psi(z_{+}, T(x_{i})) }{\sum_{z_{k} \in A(i) \cup \mathbf{C}} \psi(z_{k}, T(x_{i}))},\label{eq:center_rebalance}
\end{equation}
\text{where}
\begin{equation}
\psi(z_{k}, T(x_{i}))\!=\!\left\{
\begin{array}{ll}
\exp(z_{k} \cdot \mathcal{G}(x_{i})), & z_{k} \!\in\! A(i);   \\
\exp(z_{k} \cdot \mathcal{F}(x_{i})) \cdot q(y_{k}), & z_{k} \!\in\! \mathbf{C}. \\
\end{array}
\right.
\nonumber
\end{equation}
We emphasize that Balanced Softmax is only a practical remedy for center learning rebalance. Theoretical analysis remains future work.

\section{Experiments}
\subsection{Ablation Study}
\paragraph{Data augmentation strategy for PaCo.}
\label{Sec:data_augmentation}
Data augmentation is the key for success of contrastive learning as indicated by Chen \etal.~\cite{DBLP:conf/icml/ChenK0H20}. For PaCo, we also conduct ablation studies for different augmentation strategies. Several observations are intriguingly different from those of \cite{ouyang2016factors}.
We experiment with the following ways of data augmentation.
\begin{itemize}
	\item SimAugment: an augmentation policy \cite{DBLP:conf/cvpr/He0WXG20, DBLP:journals/corr/abs-2003-04297} that applies random flips and color jitters followed by Gaussian blur.
	\item RandAugment \cite{DBLP:conf/nips/CubukZS020}: A two stage augmentation policy that uses random parameters in place of parameters tuned by AutoAugment. The random parameters do not need to be tuned and hence reduces the search space.
\end{itemize} 

For the common {\it random resized crop} used along with the above two strategies, work of \cite{ouyang2016factors} explains that the optimal hyper-parameter for random resized crop is (0.2,1) in self-supervised contrastive learning. This setting is also adopted by work of \cite{DBLP:conf/icml/ChenK0H20, DBLP:conf/cvpr/He0WXG20, DBLP:journals/corr/abs-2003-04297, DBLP:conf/nips/GrillSATRBDPGAP20, DBLP:conf/nips/CaronMMGBJ20}. 

However, in this paper, we observe severe performance degradation on ImageNet-LT with ResNet-50 (55.0\% vs 52.2\%) for PaCo when we change the hyper-parameter from (0.08,1) to (0.2, 1).
This is because PaCo involves center learning while other self-supervised frameworks only apply non-parametric contrastive loss as described in Section \ref{Sec:contrastive_learning}.
Note that the same phenomenon is also observed on traditional supervised learning with cross-entropy loss.

Another observation is on RandAugment \cite{DBLP:conf/nips/CubukZS020}. The work of \cite{wang2021contrastive} demonstrated that directly applying strong data augmentation in MoCo \cite{DBLP:conf/cvpr/He0WXG20, DBLP:journals/corr/abs-2003-04297} does not work well. Here we observe a similar conclusion with RandAugment \cite{DBLP:conf/nips/CubukZS020}. We experiment with 3 different augmentation strategies of (1) encoder and momentum encoder input both with SimAugment; (2) encoder and momentum encoder input both with RandAugment; and (3) encoder input using RandAugment and momentum encoder input with SimAugment. The experimental results are presented in Table~\ref{tab:ablation_strongaug}. With strategy (3), PaCo achieves the best performance, showing center learning requires more aggressive data augmentation compared with contrastive learning among samples.

\begin{table}[t]
	\centering
	\caption{Comparison with different augmentation strategies for PaCo on ImageNet-LT with ResNet-50.}
	\label{tab:ablation_strongaug}
	{
		\begin{tabular}{lcc}
			\toprule
			Methods  &Aug. strategy &Top-1 Accuracy \\
			\midrule
			PaCo     &Strategy (1)   &55.0 \\
			PaCo     &Strategy (2)  &56.5 \\
			PaCo     &Strategy (3) &57.0 \\
			\bottomrule
		\end{tabular}
	}
\end{table}

\subsection{Baseline Implementation}
\label{Sec:baseline_implementation}
Contrastive learning benefits from longer training compared with traditional supervised learning with cross-entropy as Chen \etal~\cite{DBLP:conf/icml/ChenK0H20} concluded, which is also validated by previous work of \cite{DBLP:conf/icml/ChenK0H20, DBLP:conf/cvpr/He0WXG20, DBLP:journals/corr/abs-2003-04297, DBLP:conf/nips/GrillSATRBDPGAP20, DBLP:conf/nips/CaronMMGBJ20}. We run PaCo with 400 epochs on CIFAR-LT, ImageNet-LT, iNaturalist 2018, full CIFAR, and full ImageNet except for Places-LT. With Places-LT, we follow previous work \cite{Liu_2019_CVPR, DBLP:journals/corr/abs-2101-10633} by loading the pre-trained model on ImageNet and finely tune 30 epochs on Places-LT.
For fair comparison on ImageNet-LT and iNaturalist 2018, we reimplement baselines with the same training time and RandAugment \cite{DBLP:conf/nips/CubukZS020}. Especially, for RIDE, based on model ensemble, we compare with it under comparable inference latency in Fig.~\ref{fig:PaCo_comparison}.

\subsection{Long-tailed Recognition}
We follow the common evaluation protocol \cite{Liu_2019_CVPR, DBLP:journals/corr/abs-2101-10633,DBLP:conf/iclr/KangXRYGFK20} in long-tailed recognition -- that is, training models on the long-tailed source label distribution and evaluating their performance on the uniform target label distribution.
We conduct experiments on long-tailed version of CIFAR-100 \cite{cb-focal, DBLP:conf/nips/CaoWGAM19}, Places~\cite{Liu_2019_CVPR}, ImageNet \cite{Liu_2019_CVPR} and iNaturalist 2018 \cite{van2018inaturalist} datasets.

\paragraph{CIFAR-100-LT datasets.}
CIFAR-100 has 60,000 images -- 50,000 for training and 10,000 for validation with 100 categories. For fair comparison, we use the long-tailed version of CIFAR datasets with the same setting as those used in \cite{cao2019learning, zhou2019bbn, cb-focal}. They control the degrees of data imbalance with an imbalance factor $\beta$. $\beta$= $\frac{N_{max}}{N_{min}}$ where $N_{max}$ and $N_{min}$ are the numbers of training samples for the most and least frequent classes respectively. Following \cite{zhou2019bbn}, we conduct experiments with imbalance factors 100, 50, and 10.

\paragraph{ImageNet-LT and Places-LT.}
ImageNet-LT and Places-LT were proposed in \cite{Liu_2019_CVPR}. ImageNet-LT is a long-tailed version of ImageNet dataset \cite{imagenet} by sampling a subset following the Pareto distribution with power value $\alpha$=6. It contains 115.8K images from 1,000 categories, with class cardinality ranging from 5 to 1,280. Places-LT is a long-tailed version of the large-scale scene classification dataset Places \cite{zhou2017places}. It consists of 184.5K images from 365 categories with class cardinality ranging from 5 to 4,980.

\paragraph{iNaturalist 2018.}
The iNaturalist 2018 \cite{van2018inaturalist} is one species classification dataset, which is on a large scale and suffers from extremely imbalanced label distribution. It is composed of 437.5K images from 8,142 categories. In addition to the extreme imbalance, the iNaturalist 2018 dataset also confronts the fine-grained problem \cite{wei2019piecewise}.

\paragraph{Implementation details.}
For image classification on ImageNet-LT, we used ResNet-50, ResNeXt-50-32x4d, and ResNeXt-101-32x4d as our backbones for experiments. For iNaturalist 2018, we conduct experiments with ResNet-50 and ResNet-152. All models were trained using SGD optimizer with momentum $\mu = 0.9$. Contrastive learning usually requires long training time to converge. MoCo \cite{DBLP:conf/cvpr/He0WXG20, DBLP:journals/corr/abs-2003-04297}, BYOL \cite{DBLP:conf/nips/GrillSATRBDPGAP20} and SWAV \cite{DBLP:conf/nips/CaronMMGBJ20} train 800 epochs for model convergence. Supervised contrastive learning \cite{DBLP:conf/nips/KhoslaTWSTIMLK20} trains 350 epochs for feature learning and another 350 epochs for classifier learning. 

Following MoCo \cite{DBLP:conf/cvpr/He0WXG20, DBLP:journals/corr/abs-2003-04297}, when we train models with PaCo, the learning rate decays by a cosine scheduler from 0.02 to 0 with batch size 128 on 4 GPUs in 400 epochs. The temperature is set to 0.2. $\alpha$ is 0.05.
For fair comparison, we reimplement baselines in the same setting for recent state-of-the-arts of Decouple \cite{DBLP:conf/iclr/KangXRYGFK20}, Balanced Softmax \cite{DBLP:conf/nips/RenYSMZYL20} and RIDE~\cite{DBLP:journals/corr/abs-2010-01809} as mentioned in Section \ref{Sec:baseline_implementation}.

For Places-LT, following previous setting \cite{Liu_2019_CVPR, DBLP:journals/corr/abs-2101-10633},
we choose ResNet-152 as the backbone network, pre-train it on the full ImageNet-2012 dataset (provided by torchvision), and finely tune it for 30 epochs on Places-LT. Same as that on ImageNet-LT, the learning rate decays by a cosine scheduler from 0.02 to 0 with batch size 128. The temperature is set to 0.2. $\alpha$ is 0.05. For CIFAR-100-LT, we strictly follow the setting of \cite{DBLP:conf/nips/RenYSMZYL20} for fair comparison. A smaller temperature of 0.05 and $\alpha= 0.02$ are adopted for PaCo.

\begin{table}[t]
	\centering
	\caption{Long-tail recognition accuracy on ImageNet-LT for different backbone architectures.  \dag~denotes models trained with RandAugment \cite{DBLP:conf/nips/CubukZS020} in 400 epochs. More comparisons with RIDE \cite{DBLP:journals/corr/abs-2010-01809} are in Fig.~\ref{fig:PaCo_comparison}.}\small
	\label{tab:imagenet_main}
	{
		\begin{tabular}{lc@{\ \ }c@{\ \ }c} 
			\toprule
			Method & ResNet-50 & ResNeXt-50 & ResNeXt-101 \\%S & M & L  \\
			\midrule
			CE(baseline)                &41.6 &44.4 &44.8 \\
			Decouple-cRT                &47.3 &49.6 &49.4 \\
			Decouple-$\tau$-norm        &46.7 &49.4 &49.6 \\
			De-confound-TDE    &51.7 &51.8 &53.3 \\
			ResLT              &-    &52.9 &54.1 \\
			MiSLAS             &52.7 &-    & -   \\
			\midrule
			Decouple-$\tau$-norm \dag     &54.5 &56.0 &57.9 \\
			Balanced Softmax \dag         &55.0 &56.2 &58.0 \\
			PaCo\dag                      &\textbf{57.0} &\textbf{58.2} &\textbf{60.0} \\
			\bottomrule
		\end{tabular}
	}
	
\end{table}

\paragraph{Comparison on ImageNet-LT.}
Table~\ref{tab:imagenet_main} shows extensive experimental results for comparison with recent SOTA methods. We observe that Balanced Softmax \cite{DBLP:conf/nips/RenYSMZYL20} still achieves comparable results with Decouple \cite{DBLP:conf/iclr/KangXRYGFK20} across various backbones under such strong training setting on ImageNet-LT, consistent with what is claimed in the original paper. For RIDE that is based on model ensemble, we analyze the real inference speed by calculating inference time with a batch of 64 images on Nvidia GeForce 2080Ti GPU. 

We observe RIDEResNet with 3 experts even has higher inference latency than a standard ResNeXt-50-32x4d (\textbf{15.3ms vs 13.1ms}); RIDEResNeXt with 3 experts yields higher inference latency than a standard ResNeXt-101-32x4d (\textbf{26ms vs 25ms}). This result is in accordance with the conclusion that network fragmentation reduces the degree of parallelism and thus decreases efficiency in \cite{DBLP:conf/eccv/MaZZS18, DBLP:conf/iccv/CuiCLLSJ19}. For fair comparison, we do not apply knowledge distillation tricks for all these methods. As shown in Fig.~\ref{fig:PaCo_comparison} and Table~\ref{tab:imagenet_main}, under comparable inference latency, PaCo significantly surpasses these baselines. 

\paragraph{Comparison on Places-LT.}
The experimental results on Places-LT are summarized in Table~\ref{tab:place_main}. Due to the architecture change of RIDE, it is not applicable to load the publicly pre-trained model on full ImageNet, while PaCo is more flexible where the network architecture is the same as those of \cite{Liu_2019_CVPR, DBLP:journals/corr/abs-2101-10633}. Under fair training setting by finely tuning 30 epochs without RandAugment, PaCo surpasses SOTA Balanced Softmax by 2.6\%. An interesting observation is that RandAugment has little effect on the Places-LT dataset. A similar phenomenon can be observed on the iNaturalist 2018 dataset. More evaluation numbers are in the supplementary file. They can be intuitively understood since RandAugment is designed for ImageNet classification, which inspires us to explore general augmentations across different domains.

\begin{table}[t]
	\centering
	\caption{Performance on Places-LT~\cite{Liu_2019_CVPR}, starting from an ImageNet pre-trained ResNet-152 provided by torchvision. \dag denotes the model trained with RandAugment \cite{DBLP:conf/nips/CubukZS020}.}
	\label{tab:place_main}
	{
		\begin{tabular}{l|ccc|c}
			\toprule
			Method & Many & Medium & Few & \textbf{All}      \\
			\midrule
			CE(baseline)      & 45.7 & 27.3 & 8.2 & 30.2     \\
			OLTR              & 44.7 & 37.0 & 25.3 & 35.9    \\
			Decouple-$\tau$-norm & 37.8 & 40.7 & 31.8 & 37.9 \\
			Balanced Softmax     & 42.0 & 39.3 & 30.5 & 38.6 \\
			ResLT                & 39.8 & 43.6 & 31.4 & 39.8 \\
			MiSLAS               & 39.6 & 43.3 & 36.1 & 40.4 \\
			RIDE (2 experts)     & -    & -    & -    & -    \\ 
			\midrule
			PaCo                  & 37.5 & 47.2 & 33.9 & \textbf{41.2} \\
			PaCo \dag             & 36.1 & 47.9 & 35.3 & \textbf{41.2} \\      
			\bottomrule
		\end{tabular}
	}
	
\end{table}

\vspace{-.1in}
\paragraph{Comparison on iNaturalist 2018.}
Table~\ref{tab:inat_main} lists experimental results on iNaturalist 2018. Under fair training setting, PaCo consistently surpasses recent SOTA methods of Decouple, Balanced Softmax and RIDE. Our method is 1.4\% higher than Balanced Softmax. We also apply PaCo on large ResNet-152 architecture. And the performance boosts to \textbf{75.3\%} top-1 accuracy. Note that we only transfer the hyper-parameters of PaCo on ImageNet-LT to iNaturalist 2018 without any change. Tuning hyper-parameters for PaCo will bring further improvement.

\begin{table}
	\centering
	\caption{Top-1 accuracy over all classes on iNaturalist 2018 with ResNet-50. Knowledge distillation is not applied to all methods for fair comparison. We compare with RIDE {\protect\footnotemark} under comparable inference latency. \dag~denotes models trained with RandAugment \cite{DBLP:conf/nips/CubukZS020} in 400 epochs.}
	\label{tab:inat_main}
	\resizebox{.66\linewidth}{!}
	{
		\begin{tabular}{lc}
			\toprule
			Method & Top-1 Accuracy \\
			\midrule
			CB-Focal & 61.1 \\
			LDAM+DRW                   & 68.0 \\
			Decouple-$\tau$-norm       & 69.3 \\
			Decouple-LWS               & 69.5 \\
			BBN                        & 69.6 \\
			ResLT                      & 70.2 \\
			MiSLAS                     & 71.6 \\
			\midrule
			RIDE (2 experts) \dag         & 69.5  \\
			Decouple-$\tau$-norm \dag     & 71.5 \\
			Balanced Softmax \dag         & 71.8 \\
			PaCo \dag                     & \textbf{73.2} \\
			\bottomrule
		\end{tabular}
	}
\end{table}

\vspace{-.1in}
\paragraph{Comparison on CIFAR-100-LT.}
The experimental results on CIFAR-100-LT are listed in Table~\ref{tab:cifar100_main}. For the CIFAR-100-LT dataset, we mainly compare with the SOTA method Balanced Softmax \cite{DBLP:conf/nips/RenYSMZYL20} with the same training setting where Cutout \cite{DBLP:journals/corr/abs-1708-04552} and AutoAugment \cite{DBLP:conf/cvpr/CubukZMVL19} are used in training. As shown in Table~\ref{tab:cifar100_main}, PaCo consistently outperforms Balanced Softmax across different imbalance factors with such a strong setting. Specifically, PaCo surpasses Balanced Softmax by 1.2\%, 1.8\% and 1.2\% under imbalance factor 100, 50 and 10 respectively, which testify the effectiveness of our PaCo method.

\begin{table}[t]
	%   \small
	\centering
	\caption{Top-1 accuracy on CIFAR-100-LT with different imbalance factors (\dag: models trained in same setting).}
	\resizebox{.68\linewidth}{!}
	{
		\begin{tabular}{l  c  c  c}
			\toprule
			Dataset & \multicolumn{3}{c}{CIFAR-100 LT} \\ 
			\midrule
			Imbalance factor &100 &50 & 10 \\
			\midrule
			Focal Loss  & 38.4 & 44.3 & 55.8 \\
			LDAM+DRW    & 42.0 & 46.6 & 58.7 \\
			BBN         & 42.6 & 47.0 & 59.1 \\
			Causal Norm & 44.1 & 50.3 & 59.6 \\
			ResLT       & 45.3 & 50.0 & 60.8 \\
			MiSLAS      & 47.0 & 52.3 & 63.2 \\
			\midrule
			
			Balanced Softmax \dag &50.8 &54.2 &63.0 \\
			PaCo \dag    &\textbf{52.0} &\textbf{56.0} &\textbf{64.2} \\
			\bottomrule
		\end{tabular}
	}
	\label{tab:cifar100_main}
\end{table}

\subsection{Full ImageNet and CIFAR Recognition}
As analyzed in Section \ref{sec:PaCo_balance}, for balanced datasets, PaCo reduces to an improved version of multi-task learning, which adaptively adjusts the intensity of supervised cross-entropy loss and supervised contrastive loss. To verify the effectiveness of PaCo under this balanced setting, we conduct experiments on full ImageNet and full CIFAR. They are indicative to compare PaCo with supervised contrastive learning \cite{DBLP:conf/nips/KhoslaTWSTIMLK20}. Note that, under full ImageNet and CIFAR, we remove the rebalance in center learning, {\it i.e.}, Balanced Softmax.

\paragraph{Full ImageNet.}
In the implementation, we transfer hyper-parameters of PaCo on ImageNet-LT to full ImageNet without modification. SGD optimizer with momentum $\mu = 0.9$ is used. $\alpha$=0.05, temperature is 0.2 and queue size is 8,192. For multi-task training, the supervised contrastive loss is additional regularization and the loss weight is also set to 0.05.
The same data augmentation strategy is applied as PaCo, which is discussed in Section~\ref{Sec:data_augmentation}. 

The experimental results are summarized in Table~\ref{tab:full_imagenet}. With SimAugment, our ResNet-50 model achieves 78.7\% top-1 accuracy, which outperforms supervised contrastive learning model by 0.8\%. Equipped with strong augmentation, {\it i.e.}, RandAugment \cite{DBLP:conf/nips/CubukZS020}, the performance further improves to 79.3\%. ResNet-101/200 trained with PaCo consistently surpass supervised contrastive learning.

\paragraph{Full CIFAR-100.}
For CIFAR implementation, we follow supervised contrastive learning and train ResNet-50 with only the SimAugment. Compared with full ImageNet, we adopt a smaller temperature of 0.07, $\alpha=0.008$ and batch size 256 with learning rate 0.1. As shown in Table~\ref{tab:full_cifar}, on CIFAR-100, PaCo outperforms supervised contrastive learning by 2.6\%, which validates the advantages of PaCo. Note that, following \cite{DBLP:conf/iccv/CuiCLLSJ19}, we use a weight-decay of $5e-4$.

\begin{table}
	\centering
	\caption{Top-1 accuracy on full ImageNet with ResNets. ``$\star$'' denotes supervised contrastive learning with additional operation of image warping before Gaussian blur.}
	\label{tab:full_imagenet}
	\resizebox{1.00\linewidth}{!}
	{
		\begin{tabular}{lccc}
			\toprule
			Method & Model & augmentation &Top-1 Acc \\
			\midrule
			Supcon     &ResNet-50  &SimAugment $\star$   &77.9 \\
			Supcon     &ResNet-50  &RandAugment          &78.4 \\
			Supcon     &ResNet-101 &StackedRandAugment   &80.2 \\
			\midrule
			multi-task &RandAugment         &ResNet-50 &78.1 \\
			\midrule
			PaCo        &ResNet-50  &SimAugment           &\textbf{78.7} \\
			PaCo        &ResNet-50  &RandAugment          &\textbf{79.3} \\
			PaCo        &ResNet-101 &StackedRandAugment   &\textbf{80.9} \\
			PaCo        &ResNet-200 &StackedRandAugment   &\textbf{81.8} \\
			\bottomrule
		\end{tabular}
	}
\end{table}

\begin{table}
	\centering
	\caption{Top-1 accuracy on full CIFAR-100 (ResNet-50).}
	\label{tab:full_cifar}
	\resizebox{.76\linewidth}{!}
	{
		\begin{tabular}{lccc}
			\toprule
			Method & dataset &Top-1 Accuracy \\
			\midrule
			CE(baseline)    &CIFAR-100    &77.9 \\
			multi-task      &CIFAR-100    &78.0 \\
			\midrule
			Supcon          &CIFAR-100    &76.5 \\
			PaCo             &CIFAR-100    &\textbf{79.1} \\
			\bottomrule
		\end{tabular}
	}
\end{table}

\section{Conclusion}
\vspace{-0.1in}
In this paper, we have proposed Parametric Contrastive Learning (PaCo), which contains a set of parametric class-wise learnable centers to tackle the long-tailed recognition. It is based on the theoretical analysis of supervised contrastive learning. For balanced data, our analysis of PaCo demonstrates that it can adaptively enhance the intensity of pushing two samples of the same class close as more samples are pulled together with their corresponding centers, which can potentially benefit hard examples learning in training. 

We conduct experiments on various benchmarks of CIFAR-LT, ImageNet-LT, Places-LT, and iNaturalist 2018. The experimental results show that we create a new state-of-the-art for long-tailed recognition. Further, experimental results on full ImageNet and CIFAR demonstrate that PaCo also benefits balanced datasets.

{\small
	\bibliographystyle{ieee_fullname}
	\bibliography{egbib}
}

\newpage
\onecolumn
\appendix

\begin{center}
	\Large \textbf{Parametric Contrastive Learning}
	\Large \\ \textbf{Supplementary Material}
\end{center}
\vspace{20pt}

\section{Proof to Remark 1}
For an image $X_{i}$ and its label $y_{i}$, the expectation number of positive pairs with respect to $X_{i}$ will be: 

\begin{equation}
K_{y_{i}} = q(y_{i}) * (\mathtt{length}(queue) + batchsize * 2 - 1) \approx \mathtt{length}(queue) \cdot q(y_{i}), 
\label{eq:K_approx}
\end{equation}

$q(y_{i})$ is the class frequency over the whole dataset. Here the "$\approx$" establishes because $batchsize \ll \mathtt{length}(queue)$ in training process. Note that we use such approximation just for simplification. Our analysis holds for the precise $K_{y_{i}}$. In what follows, we prove the optimal values for supervised contrastive loss.

Suppose training samples are i.i.d.
To minimize the supervised contrastive loss for sample $X_{i}$, according to Eq.~\eqref{eq:supcon}, we rewrite:
\begin{equation}
\left\{
\begin{aligned}
P(i)&=&\{z_{1}^{+}, z_{2}^{+}, ..., z_{K_{y_{i}}}^{+} \}; \nonumber \\
p_{i}^{+}&=&\frac{ \exp({z_{i}^{+} \cdot T(x_{i})}) }{\sum_{z_{k} \in A(i)} \exp({z_{k} \cdot T(x_{i})}) }; \nonumber \\
p_{sum}^{+} &=& p_{1}^{+} + p_{2}^{+}+...+p_{K_{y_{i}}}^{+}. \nonumber \\
\end{aligned}
\right.
\end{equation}

Then the supervised contrastive loss will be:
\begin{eqnarray}
\mathcal{L}_{i} \!&=&\! -\sum_{ z_{+} \in P(i)} \log \frac{ \exp({z_{+} \cdot T(x_{i})}) }{\sum_{z_{k} \in A(i)} \exp({z_{k} \cdot T(x_{i})})}  \nonumber \\
&=& - (\log p_{1}^{+} + \log p_{2}^{+} + ... + \log p_{K_{y_{i}}}^{+}). \nonumber
\end{eqnarray}

For obtaining its optimal solution, we define the Lagrange multiplier form of $\mathcal{L}_{i}$ as:

\begin{equation}
l = - (\log p_{1}^{+} + \log p_{2}^{+} + ... + \log p_{K_{y_{i}}}^{+}) + \lambda (p_{1}^{+} + p_{2}^{+}+...+p_{K_{y_{i}}}^{+} - p_{sum}^{+}), \label{eq:largrange}
\end{equation}

where $\lambda$ is the Lagrange multiplier. The first order conditions of Eq.~\eqref{eq:largrange} w.r.t. $\lambda$ and $p_{i}^{+}$ can be written as follows:
\begin{equation}
\left\{
\begin{aligned}
\frac{\partial l}{\partial p_{i}^{+}} &= -\frac{1}{p_{i}^{+}} + \lambda = 0;\\
\frac{\partial l}{\partial \lambda} &= p_{1}^{+} + p_{2}^{+}+...+p_{K_{y_{i}}}^{+} - p_{sum}^{+} =0. \\
\end{aligned}
\right.
\label{eq:largrange_conditions}
\end{equation}

From Eq.~\eqref{eq:largrange_conditions}, the optimal solution for $p_{i}^{*}$ will be $\cfrac{p_{sum}^{+}}{K_{y_{i}}}$. Note that $p_{sum}^{+} \in [0,1]$,  with a specific $p_{sum}^{+}$, the minimal loss value of $\mathcal{L}_{i}$ is:
\begin{equation}
\mathcal{L}_{i} = -K_{y_{i}} \log \frac{p_{sum}^{+}}{K_{y_{i}}}.
\end{equation}

Thus, when $p_{sum}^{+}=1.0$, $\mathcal{L}_{i}$ achieves minimum with the optimal value $p_{i}^{+}=\cfrac{1}{K_{y_{i}}}$ which is exactly the probability that two samples of the same class are a true positive pair.

\newpage        
\section{Proof to Remark 2}
For the image $X_{i}$ and its label $y_{i}$,  Eq.~\eqref{eq:K_approx} still establishes for our parametric contrastive loss. To minimize the parametric contrastive loss for sample $X_{i}$, according to Eq. \eqref{eq:PaCo}, we similarly rewrite:

\begin{equation}
\left\{
\begin{aligned}
P(i)&=&\{z_{1}^{+}, z_{2}^{+}, ..., z_{K_{y_{i}}}^{+} \} \nonumber \\
p_{i}^{+}&=&\frac{ \exp({z_{i}^{+} \cdot T(x_{i})}) }{\sum_{z_{k} \in A(i) \cup \mathbf{C}} \exp({z_{k} \cdot T(x_{i})})} \nonumber \\
p_{c}^{+}&=&\frac{exp(c_{y} \cdot T(x_{i}))}{\sum_{z_{k} \in A(i) \cup \mathbf{C}} \exp({z_{k} \cdot T(x_{i})})} \\
p_{sum}^{+} &=& p_{1}^{+} + p_{2}^{+}+...+p_{K_{y_{i}}}^{+} + p_{c}^{+}. \nonumber \\
\end{aligned}
\right.
\end{equation}

Then the parametric contrastive loss will be:

\begin{eqnarray}
\mathcal{L}_{i} \!&=&\! \sum_{ z_{+} \in P(i) \cup \{c_{y}\} } \!-w(z_{+}) \log \frac{ \exp({z_{+} \cdot T(x_{i})}) }{\sum_{z_{k} \in A(i) \cup \mathbf{C}} \exp({z_{k} \cdot T(x_{i})})} \\
&=& -\left( \log p_{c}^{+} +  \alpha \cdot (\log p_{1}^{+} + \log p_{2}^{+} + ... + \log p_{k_{y_{i}}}^{+} )     \right).
\end{eqnarray}

For obtaining its optimal solution, we define the Lagrange multiplier form of $\mathcal{L}_{i}$ as:

\begin{equation}
l = -\left( \log p_{c}^{+} +  \alpha \cdot (\log p_{1}^{+} + \log p_{2}^{+} + ... + \log p_{k_{y_{i}}}^{+} )     \right) + \lambda (p_{1}^{+} + p_{2}^{+}+...+p_{K_{y_{i}}}^{+} + p_{c}^{+} - p_{sum}^{+}), \label{eq:largrange_paco}
\end{equation}

where $\lambda$ is the Lagrange multiplier. The first order conditions of Eq.~\eqref{eq:largrange_paco} w.r.t. $\lambda$, $p_{c}^{+}$ and $p_{i}^{+}$ can be written as follows:

\begin{equation}
\left\{
\begin{aligned}
\frac{\partial l}{\partial p_{i}^{+}} &= - \frac{\alpha}{p_{i}^{+}} + \lambda = 0; \\
\frac{\partial l}{\partial p_{c}^{+}} &= -\frac{1}{p_{c}^{+}} + \lambda = 0; \\
\frac{\partial l}{\partial \lambda} &= p_{1}^{+} + p_{2}^{+}+...+p_{K_{y_{i}}}^{+} + p_{c}^{+} - p_{sum}^{+} =0. \\
\end{aligned}
\right.
\label{eq:largrange_conditions_paco}
\end{equation}

From Eq.~\eqref{eq:largrange_conditions_paco}, the optimal solution for $p_{i}^{+}$ and $p_{c}^{+}$ will be $\cfrac{\alpha p_{sum}^{+}}{1+ \alpha K_{y_{i}}}$ and $\cfrac{p_{sum}^{+}}{1+ \alpha K_{y_{i}}}$ respectively. Note that $p_{sum}^{+} \in [0,1]$,  with a specific $p_{sum}^{+}$, the minimal loss value of $\mathcal{L}_{i}$ is:

\begin{equation}
\mathcal{L}_{i} = - \log \cfrac{p_{sum}^{+}}{1+ \alpha K_{y_{i}}} - \alpha K_{y_{i}} \log \frac{\alpha p_{sum}^{+}}{1+\alpha K_{y_{i}}}. 
\end{equation}

Thus, when $p_{sum}^{+}=1.0$, $\mathcal{L}_{i}$ achieves minimum with the optimal value $p_{i}^{+}=\cfrac{\alpha}{1+\alpha K_{y_{i}}}$ , which is the probability that two samples of the same class are a true positive pair, and the optimal value $p_{c}^{+}=\cfrac{1}{1+\alpha K_{y_{i}}}$ which is the probability that a sample is closest to its corresponding center $c_{y_{i}}$ among $\mathbf{C}$.

\newpage
\section{Gradient Derivation}
In Section~\ref{sec:PaCo_balance}, we analyze PaCo loss under balanced setting, taking full ImageNet as an example. With $P_{sup}$ increases from 0 to 0.71, the intensity of supervised contrastive loss will enlarge. Here we show that more samples will be pulled together with their corresponding centers when $P_{sup}$ increases from 0 to 0.71 from the perspective of gradient derivation.

\begin{equation}
\frac{\partial \mathcal{L}}{\partial c_{k}} = \left \{
\begin{aligned}
&(\alpha K^{*} +1 )p_{c_{k}} x_{i}, &\quad y_{i} \neq k; \\
&\left\{ (\alpha K^{*} + 1)p_{c_{k}} -1 \right\} x_{i}, &\quad y_{i}=k. \\
\end{aligned}
\right.	
\label{eq:gradient_derivation}
\end{equation}

It is worthy to note that when $p_{c_{k}} \in (0,0.71)$, we have
\begin{equation}
\left \{
\begin{aligned}
&\frac{\partial \mathcal{L}}{\partial c_{k}} >0, &\quad y_{i} \neq k; \\
&\frac{\partial \mathcal{L}}{\partial c_{k}} <0, &\quad y_{i}=k. \\
\end{aligned}
\right.	
\label{eq:gradient_derivation2}
\end{equation}
Eqs.~\eqref{eq:gradient_derivation} and \eqref{eq:gradient_derivation2} mean that as $P_{sup}$ increases in training process, the probability that a sample is closest to its corresponding center will increase and the probability that a sample is closest to other centers will decrease. Thus, more and more samples will be pulled together with their right centers.

\section{More Experimental Results on Many-shot, Medium-shot, and Few-shot.}

\begin{table*}[h]
	\caption{Comprehensive results on ImageNet-LT with different backbone networks (ResNet-50, ResNeXt-50 \& ResNeXt-101). Models are trained with RandAugment in 400 epochs. Inference time is calculated with a batch of 64 images on Nvidia GeForce 2080Ti GPU, Pytorch1.5, Python3.6.}
	\label{tab:imagenet_extra}
	%\vspace{-0.2in}
	\begin{center}
		\begin{tabular}{llccccc}
			\toprule
			Backbone & Method & \textbf{Inference time (ms)} & Many & Medium & Few & All\\
			\midrule
			\multirow{3}{*}{ResNet-50}
			&$\tau$-normalize &8.3 &65.0 &52.2 &32.3 &54.5 \\
			&Balanced Softmax &8.3 &66.7 &52.9 &33.0 &55.0 \\
			&PaCo             &8.3 &65.0 &55.7 &38.2 &57.0\\
			\midrule
			\multirow{3}{*}{ResNeXt-50}
			&$\tau$-normalize &13.1 &66.4 &53.4 &38.2 &56.0 \\
			&Balanced Softmax &13.1 &67.7 &53.8 &34.2 &56.2\\
			&PaCo             &13.1 &67.5 &56.9 &36.7 &58.2\\
			\midrule
			\multirow{3}{*}{ResNeXt-101}
			&$\tau$-normalize &25.0 &69.0 &55.1 &36.9 &57.9 \\
			&Balanced Softmax &25.0 &69.2 &55.8 &36.3 &58.0 \\
			&PaCo             &25.0 &68.2 &58.7 &41.0 &60.0 \\
			\bottomrule
		\end{tabular}
	\end{center}
\end{table*}

\begin{table*}[h]
	\caption{Comprehensive results on ImageNet-LT with RIDE. Models are trained with RandAugment in 400 epochs. Inference time is calculated with a batch of 64 images on Nvidia GeForce 2080Ti GPU, Pytorch1.5, Python3.6.}
	%\vspace{-0.2in}
	\label{tab:imagenet_ride}
	\begin{center}
		\begin{tabular}{llccccc}
			\toprule
			Backbone & Method & \textbf{Inference time (ms)} & Many & Medium & Few & All\\
			\midrule
			\multirow{3}{*}{RIDEResNet}
			&1 expert  &8.2  &64.8 &49.8 &29.6 &52.8 \\
			&2 experts &12.0 &67.7 &53.5 &31.5 &56.0 \\
			&3 experts &15.3 &69.0 &54.7 &32.5 &57.0 \\
			\midrule
			\multirow{3}{*}{RIDEResNeXt}
			&1 expert  &13.0 &67.2 &49.0 &28.1 &53.2 \\
			&2 experts &19.0 &70.4 &52.6 &30.3 &56.4 \\
			&3 experts &26.0 &71.8 &53.9 &32.0 &57.8 \\
			\bottomrule
		\end{tabular}
	\end{center}
\end{table*}

\begin{table*}[h]
	\caption{Comprehensive results on iNaturalist 2018 with ResNet-50 and ResNet-152. \dag represents the models are trained without RandAugment. Inference time is calculated with a batch of 64 images on Nvidia GeForce 2080Ti GPU, Pytorch1.5, Python3.6.}
	%\vspace{0.2in}
	\label{tab:inat_extra}
	\begin{center}
		\begin{tabular}{llccccc}
			\toprule
			Backbone & Method & \textbf{Inference time (ms)} & Many & Medium & Few & All\\
			\midrule
			\multirow{3}{*}{ResNet-50}
			&$\tau$-normalize &8.3 &74.1 &72.1 &70.4 &71.5 \\
			&Balanced Softmax &8.3 &72.3 &72.6 &71.7 &71.8 \\
			&PaCo             &8.3 &70.3 &73.2 &73.6 &73.2 \\
			\midrule
			\multirow{2}{*}{ResNet-50 \dag}
			&Balanced Softmax &8.3 &72.5 &72.3 &71.4 &71.7 \\
			&PaCo             &8.3 &69.5 &73.4 &73.0 &73.0 \\
			\midrule
			\multirow{1}{*}{ResNet-152}         
			&PaCo &20.1 &75.0 &75.5 &74.7 &75.2  \\
			\bottomrule
		\end{tabular}
	\end{center}
\end{table*}

\newpage
\begin{table*}[t]
	\caption{Comprehensive results on iNaturalist 2018 with RIDE. Models are trained with RandAugment in 400 epochs without knowledge distillation. Inference time is calculated with a batch of 64 images on Nvidia GeForce 2080Ti GPU, Pytorch1.5, Python3.6.}
	%\vspace{-0.2in}
	\label{tab:inat_ride}
	\begin{center}
		\begin{tabular}{llccccc}
			\toprule
			Backbone & Method & \textbf{Inference time (ms)} & Many & Medium & Few & All\\
			\midrule
			\multirow{4}{*}{RIDEResNet}
			&1 expert  &8.2  &56.0 &66.3 &66.0 &65.2 \\
			&2 experts &12.0 &62.2 &70.5 &70.0 &69.5 \\
			&3 experts &15.3 &66.5 &72.1 &71.5 &71.3 \\
			\bottomrule
		\end{tabular}
	\end{center}
\end{table*}

\section{Implementation details for Table 1}
We train models with cross-entropy, parametric contrastive loss 400 epochs without RandAugment respectively. For supervised contrastive loss, following the original paper, we firstly train the model 400 epochs. Then we fix the backbone and train a linear classifier 400 epochs. 

\section{Ablation Study}
\paragraph{Re-weighting in contrastive learning without center learning rebalance}
Re-weighting is a classical method for dealing with imbalanced data. Here we directly apply the re-weighting method of Cui \etal.~\cite{cb-focal} in contrastive learning to compare with PaCo. Moreover, Balanced softmax \cite{DBLP:conf/nips/RenYSMZYL20}, as one state-of-the-art method for traditional cross-entropy in long-tailed recognition, is also applied to contrastive learning rebalance. The experimental results are summarized in Table~\ref{tab:ablation_reweighting_supp}. It is obvious PaCo significantly surpasses the two baselines.
\begin{table}[t]
	\centering
	\caption{Comparison with re-weighting baselines on ImageNet-LT with ResNet-50. The re-weighting strategy is applied to the supervised contrastive loss. Models are all trained without RandAugment. }
	\label{tab:ablation_reweighting_supp}
	\vspace{0pt}
	{
		\begin{tabular}{lc}
			\toprule
			Method  &Top-1 Accuracy \\
			\midrule
			CE                              &48.4 \\
			multi-task (CE+Re-weighting)    &49.0 \\
			multi-task (CE+Blance Softmax)  &48.6 \\
			\midrule
			PaCo                &51.0 \\
			\bottomrule
		\end{tabular}
	}
\end{table}

\paragraph{Rebalance in center learning}
PaCo balances the contrastive learning (for moderating contrast among samples). However the center learning also needs to be balanced, which has been explored in \cite{DBLP:journals/nn/BudaMM18,DBLP:conf/cvpr/HuangLLT16,DBLP:conf/cvpr/CuiJLSB19, he2009learning,chawla2002smote, shen2016relay, DBLP:conf/nips/RenYSMZYL20, DBLP:conf/iclr/KangXRYGFK20, DBLP:journals/corr/abs-2010-01809, DBLP:journals/corr/abs-2101-10633, DBLP:conf/nips/TangHZ20,duggal2020elf}. To compare with state-of-the-art methods in long-tailed recognition, we incorporate Balanced Softmax \cite{DBLP:conf/nips/RenYSMZYL20} into the center learning. As shown in Table~\ref{tab:ablation_center_rebalance_supp}, after rebalance in center learning, PaCo boosts performance to 58.2\%, surpassing baselines with a large margin. 

\begin{table}[t]
	\centering
	\caption{Comparison with re-weighting baselines that perform center learning rebalance on ImageNet-LT with ResNeXt-50. Models are all trained with RangAugment in 400 epochs.}
	\label{tab:ablation_center_rebalance_supp}
	\vspace{0pt}
	{
		\begin{tabular}{lcc}
			\toprule
			Method  &loss weight &Top-1 Accuracy \\
			\midrule
			multi-task (Balanced Softmax \!+\! Re-weighting)      &0.05 &57.0  \\
			multi-task (Balanced Softmax \!+\! Re-weighting)      &0.10 &57.1  \\
			multi-task (Balanced Softmax \!+\! Re-weighting)      &0.20 &57.1  \\
			multi-task (Balanced Softmax \!+\! Re-weighting)      &0.30 &57.0  \\
			multi-task (Balanced Softmax \!+\! Re-weighting)      &0.50 &57.2  \\
			multi-task (Balanced Softmax \!+\! Re-weighting)      &0.80 &57.2  \\
			multi-task (Balanced Softmax \!+\! Re-weighting)      &1.00 &56.9  \\
			\midrule
			PaCo                &- &58.2 \\
			\bottomrule
		\end{tabular}
	}
\end{table}

\end{document}